\pdfoutput=1

\documentclass[11pt]{article}

\usepackage{ACL2023}

\usepackage{times}
\usepackage{latexsym}

\usepackage[T1]{fontenc}

\usepackage[utf8]{inputenc}

\usepackage{microtype}

\usepackage{inconsolata}

\usepackage{booktabs}
\usepackage{multirow}
\usepackage{enumitem}

\usepackage{amsmath}
\usepackage{amssymb}
\usepackage{graphicx}
\usepackage{subcaption}
\usepackage[export]{adjustbox}

\usepackage{xcolor}
\definecolor{NYUViolet}{RGB}{87,6,140}
\definecolor{NYULightViolet1}{RGB}{171,130,197}
\definecolor{NYUTeal}{RGB}{0,156,139}
\definecolor{NYUMagenta}{RGB}{224,15,120}

\DeclareMathOperator{\weat}{weat}
\DeclareMathOperator*{\mean}{mean}
\DeclareMathOperator*{\stdev}{stdev}

\title{Reflecting the Male Gaze: Quantifying Female Objectification in 19th and 20th Century Novels}

\author{Kexin Luo, Yue Mao, Bei Zhang, \and Sophie Hao \\
        Center for Data Science, New York University\\
    	\texttt{\{kl3108,ym1596,bz2428,sh7354\}@nyu.edu}}

\begin{document}
\maketitle
\begin{abstract}
    Inspired by the concept of the \textit{male gaze} \citep{mulveyVisualPleasureNarrative1975} in literature and media studies, this paper proposes a framework for analyzing gender bias in terms of \textit{female objectification}---the extent to which a text portrays female individuals as objects of visual pleasure. Our framework measures female objectification along two axes. First, we compute an \textit{agency bias} score that indicates whether male entities are more likely to appear in the text as grammatical agents than female entities. Next, by analyzing the word embedding space induced by a text \citep{caliskanSemanticsDerivedAutomatically2017}, we compute an \textit{appearance bias} score that indicates whether female entities are more closely associated with appearance-related words than male entities. Applying our framework to 19th and 20th century novels reveals evidence of female objectification in literature: we find that novels written from a male perspective systematically objectify female characters, while novels written from a female perspective do not exhibit statistically significant objectification of any gender.
\end{abstract}

\section{Introduction}

    In literature and media studies, the \textit{male gaze} \citep{mulveyVisualPleasureNarrative1975} refers to a phenomenon in which women are depicted in film, literature, and the visual arts as objects of aesthetic pleasure, to be consumed and enjoyed by a heterosexual male viewer. The male gaze, and the practice of \textit{female objectification} more generally, perpetuates an understanding of women as tools meant to serve the interests of others, with little attention paid to women's agency, individuality, or subjectivity \citep{nussbaumObjectification1995}. Beyond these representational harms \citep{barocasProblemBiasAllocative2017,crawfordTroubleBias2017}, there is evidence that internalization of female objectification can result in averse mental health impacts on girls and women \citep{fredricksonObjectificationTheoryUnderstanding1997,szymanskiSexualObjectificationWomen2011}. Despite its harms, however, female objectification seems to be ubiquitous in Western culture. Within the computational linguistics literature, male-gaze-like depictions of women have been documented in film and television \citep{agarwalKeyFemaleCharacters2015,singhFemaleAstronautBecause2023}, Tweets \citep{anzovinoAutomaticIdentificationClassification2018}, internet memes \citep{fersiniSemEval2022TaskMultimedia2022,singhFemaleAstronautBecause2023}, text corpora \citep{dacunhaObjectifyingWomenSyntactic2022}, text generated by language models \citep{lucyGenderRepresentationBias2021,wolfeContrastiveLanguageVisionAI2023}, and even theoretically ``gender-neutral'' media such as linguistics textbooks \citep{macaulayDonTouchMy1997} and journal articles \citep{kotekGenderRepresentationLinguistic2020}.

    This paper proposes a quantitative framework for studying female objectification in text, leveraging techniques developed for the analysis of gender bias in natural language processing (NLP). We operationalize the concept of female objectification by factoring it into two biases that a text might exhibit: an \textit{agency bias} that favors treating male entities as grammatical agents, and an \textit{appearance bias} that favors mentioning female entities in collocation with words related to appearance. To measure these biases within a collection of texts, we develop a pipeline that combines several NLP tools: we measure agency bias by using a semantic role labeler \citep{shiSimpleBERTModels2019} to analyze the argument structure of male and female entities in a text, and we measure appearance bias by using the Word Embedding Association Test \citep{caliskanSemanticsDerivedAutomatically2017} to analyze stereotypical associations in the word embedding space induced by the text.

    We apply our framework by presenting a quantitative study of female objectification in English-language novels and translations from the late modern era. We show that commonly-downloaded open-source novels exhibit positive, statistically significant levels of agency bias and appearance bias, revealing the existence of systematic female objectification within this repertoire. When controlling for the gender of authors and narrators, we find that novels with a female author or a first-person female narrator do not exhibit statistically significant levels of either agency bias or appearance bias---some are strongly biased, but most are weakly biased in the opposite direction (i.e., they exhibit mild levels of \textit{male objectification}). On the other hand, novels with a male author or a first-person male narrator consistently exhibit both agency bias and appearance bias.

    To summarize, our contributions are as follows. 
    In \autoref{sec:metrics}, we define two bias metrics for texts, which formalize the concept of female objectification. In \autoref{sec:setup}, we develop a methodology for studying female objectification in a collection of texts. Finally, in \autoref{sec:results}, we present empirical evidence of a male gaze phenomenon in 19th and 20th century English-language novels.\footnote{Our code is available at \url{https://github.com/Bellaaazzzzz/Female_Objectification_Quantifying_in_Novels}.}

    \section{Measuring Female Objectification}
    \label{sec:metrics}

    Our treatment of female objectification is inspired by the following elements of the male gaze: (1) that women are depicted as \textit{passive} objects, (2) to be consumed for \textit{visual} pleasure. We operationalize these two concepts by defining the following bias metrics. 
    \begin{enumerate}
        \item \textbf{Agency Bias:} A text exhibits \textit{agency bias} if male entities are more likely than female entities to appear in the text as grammatical agents.
        \item \textbf{Appearance Bias:} A text exhibits \textit{appearance bias} if ``female'' words are distributionally closer to ``appearance'' words than ``male'' words.
    \end{enumerate}
    For both bias metrics, a value of 0 represents lack of bias, a positive value represents presence of female objectification, and a negative value represents presence of male objectification.

    \subsection{Agency Bias}

    The \textit{grammatical agent} of a clause is the entity that initiates the event denoted by the main predicate of that clause. For example, in the sentence \textit{Bob was seen by Alice}, \textit{Alice} is the agent, since Alice initiates the act of seeing. \textit{Bob}, the person who receives the act of seeing, is the \textit{patient}. Our agency bias metric is based on the intuition that if a text portrays women as passive objects, then female entities are less likely than male entities to appear in the text as grammatical agents. 

    \subsubsection{Definition of Agency Bias}
    We define the \textit{female agentivity} (resp. \textit{male agentivity}) of a text as the conditional probability that an entity appears in the text as a grammatical agent, given that it is female (resp. male). The agency bias of a text is defined as follows:
    \[
    \text{agency bias} = \frac{\text{male agentivity}}{\text{female agentivity}} - 1\text{.}
    \]

    \begin{figure}
        \centering
        \textcolor{NYUTeal}{\textbf{Alice}} saw \textcolor{NYULightViolet1}{\textit{Bob}} at the park. \textcolor{NYUTeal}{\textbf{She}} waved to \textcolor{NYULightViolet1}{\textit{him}} and said, ``Hello!'' \textcolor{NYUViolet}{\textbf{Bob}} smiled and walked over.\vspace{1em}

        \footnotesize 
        \begin{tabular}{r r c r}
            \textbf{Male Agentivity:} & $1/3$ & $=$ & $.33$ \\
            \textbf{Female Agentivity:} & $2/2$ & $=$ & $1.00$ \\\midrule
            \textbf{Agency Bias:} & $\frac{1/3}{2/2} - 1$ & $=$ & $-.67$
        \end{tabular}
        \caption{Calculating agency bias for a short story by counting occurrences of \textcolor{NYUTeal}{\textbf{female agents}}, \textcolor{NYUViolet}{\textbf{male agents}}, and \textcolor{NYULightViolet1}{\textit{male patients}}.}
        \label{fig:agency-bias-calculation}
    \end{figure}

    \paragraph{Example.} \autoref{fig:agency-bias-calculation} illustrates how agency bias is calculated for a short story. We estimate male agentivity to be $1/3$, since there are three occurrences of a male entity (Bob), and one of them is an agent. The female agentivity of this text is $1$, since the sole female entity in this story (Alice) acts as an agent in both occurrences. The final agency bias is $-2/3$, meaning that female entities are 67\% more likely than male entites to appear in this story as agents.
 
    \subsubsection{Calculating Agency Bias}
    \label{sec:agencybias-calculation}

    To calculate agency bias for a text at scale, we use a procedure consisting of the following steps.
    \begin{itemize}
        \item \textbf{Entity Extraction:} We extract entities from the text using spaCy's named entity recognizer.
        \item \textbf{Gender Classification:} We classify these entities as \textit{male} or \textit{female} using a procedure similar to the one used in \citet{toroisazaAreFairyTales2023}.
        \item \textbf{Agency Classification:} We classify the entities as \textit{agents} or \textit{non-agents} using \citeauthor{shiSimpleBERTModels2019}'s (\citeyear{shiSimpleBERTModels2019}) semantic role labeler.
    \end{itemize}
    The full details of our implementation are given in \autoref{sec:appendix}.

    \subsection{Appearance Bias}

    The Word Embedding Association Test (WEAT, \citealp{caliskanSemanticsDerivedAutomatically2017}) quantifies the extent to which a word embedding space conveys stereotypical associations between demographic groups. Our appearance bias metric uses WEAT to compare ``male'' and ``female'' words in terms of their similarity to ``appearance'' words. If a text depicts women as bearers of visual pleasure, then we expect the female words to be closer than male words to the appearance words when an embedding space is trained on that text.

    \subsubsection{Definition of Appearance Bias}
    
    Let $M$, $F$, and $A$ be sets of \textit{male words}, \textit{female words}, and \textit{appearance words}, respectively.\footnote{Unlike \citet{caliskanSemanticsDerivedAutomatically2017}, we only use one set of target words.} Let $\mathbb{W}$ be a word embedding space, where the embedding of a word $w$ is denoted by $\vec{w}$. The \textit{WEAT score for $\mathbb{W}$} is defined as the quantity
    \[
    \weat(\mathbb{W}) = \frac{\mean_{a \in A} s(a)}{\stdev_{a \in A} s(a)}\text{,}
    \]
    where
    \[
    s(a) = \mean_{f \in F} \cos(\vec{f}, \vec{a}) - \mean_{m \in M} \cos(\vec{m}, \vec{a})\text{.}
    \]
    To compute the appearance bias of a text, we take a pre-trained embedding space $\mathbb{W}$, and fine-tune it on the text. Letting $\mathbb{W}^\prime$ denote the fine-tuned embedding space, the appearance bias of the text is defined as:
    \[
    \text{appearance bias} = \weat(\mathbb{W}^\prime) - \weat(\mathbb{W})\text{.}
    \]

    \begin{table}
        \centering
        \footnotesize
        \begin{tabular}{l | l}
            \toprule 
            \textbf{Male} & boy, brother, father, he, him,  \\
            \textbf{Words} ($M$) & himself, husband, male, man, mr,  \\
            & sir, uncle, \textcolor{NYULightViolet1}{male named entities} \\\midrule
            \textbf{Female} & aunt, female, girl, her, herself, lady,   \\
            \textbf{Words} ($F$) & miss, mother, she, sister, wife,  \\
            & woman, \textcolor{NYULightViolet1}{female named entities} \\\midrule
            \textbf{Appearance} & belt, complexion, dress, eye, lip,  \\
            \textbf{Words} ($A$) & outfit, plain, pore, purse, ravishing,  \\
            & ugly, voluptuous, \textcolor{NYULightViolet1}{992 others} \\\bottomrule
        \end{tabular}
        \caption{Words used to calculate WEAT scores for the appearance bias metric.}
        \label{table:weat-words}
    \end{table}

    \paragraph{Word Sets.} \autoref{table:weat-words} shows the male, female, and appearance words used to calculate WEAT scores on our pre-trained and fine-tuned embedding spaces. The full set of appearance words is obtained from the Oxford Learner's Dictionaries' ``Appearance'' topic vocabulary.\footnote{\url{http://www.oxfordlearnersdictionaries.com/topic/category/appearance_1}} The male and female words include those listed in \autoref{table:weat-words}, as well as named entities that can be assigned a gender.

    \begin{figure}
        \resizebox{\columnwidth}{!}{%
            \includegraphics{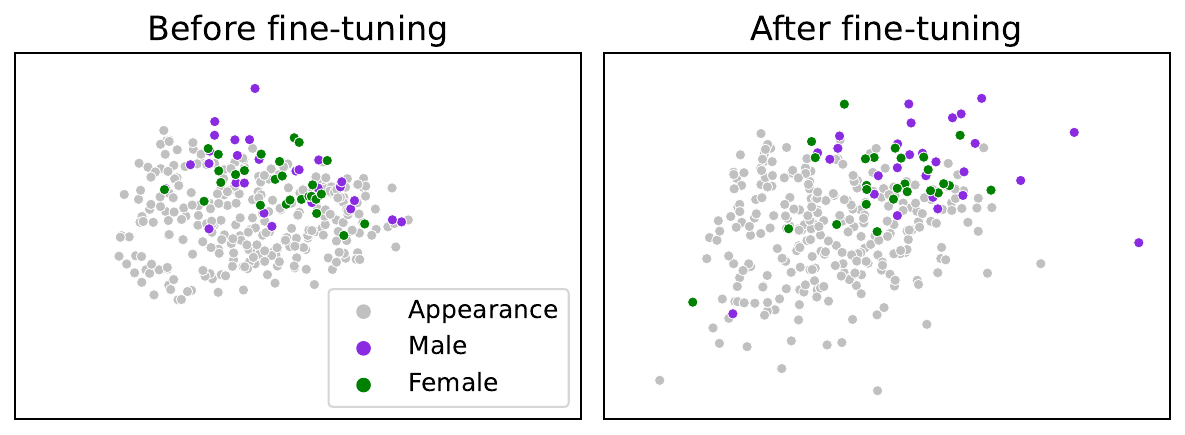}%
        } 
        \caption{\label{fig:WEAT_visualization} GloVe embeddings \citep{penningtonGloveGlobalVectors2014} for words in \autoref{table:weat-words}, plotted along two principal components. Fine-tuning the embeddings on the novel \textit{Lady Audley’s Secret} by Mary Elizabeth Braddon causes ``male'' words, but not ``female'' words, to move away from the cluster of ``appearance'' words. The novel's appearance bias is 1.575.}
    \end{figure}

    \paragraph{Example.} \autoref{fig:WEAT_visualization} visualizes how a word embedding space changes after it has been fine-tuned on a novel. In this example, the three word clusters overlap in the pre-trained embedding space, but the male words drift away from the appearance words after fine-tuning. The appearance bias of this text is 1.575, indicating that this text associates females with appearance more than males.

    \subsubsection{Calculating Appearance Bias}
    \label{sec:appearancebias-calculation}

    Here we briefly describe elements of our procedure for calculating the appearance bias of a text. Full implementation details are given in \autoref{sec:appendix}.

    \paragraph{Gendered Entities.} Similar to our agency bias pipeline, our calculation of appearance bias includes entity extraction and gender classification steps that produce the named entities included among the male and female words.

    \paragraph{Embedding Spaces.} We use \texttt{glove.6B} embeddings \citep{penningtonGloveGlobalVectors2014}, pre-trained on Wikipedia articles and the Gigaword corpus. We fine-tune our embeddings using the CBOW objective with negative sampling \citep{mikolovEfficientEstimationWord2013,mikolovDistributedRepresentationsWords2013}. Since proper nouns may be out of vocabulary, we randomly (re-)initialize the embeddings for named entities before fine-tuning. %

    \section{The Male Gaze in Literature}
    \label{sec:setup}

    Our main experiment uses our framework to quantitatively study female objectification in English-language novels and translations from the late modern period. We compile a collection of open-source novels primarily from the 19th and 20th centuries, and measure the appearance bias and agency bias of these novels. We conclude that our collection of novels exhibits evidence of systematic female objectification if we are able to reject the null hypothesis that the average appearance bias and agency bias of the novels are both 0. Our goal is to answer the following questions.
    \begin{itemize}
        \item[Q1.] Do novels exhibit systematic female objectification in general?
        \item[Q2.] Is the use of female objectification influenced by the gender of a novel's author or narrator?
    \end{itemize}

    \subsection{Texts}
    
\begin{figure}
        \centering
        \footnotesize %
        \resizebox{\columnwidth}{!}{%
            \includegraphics{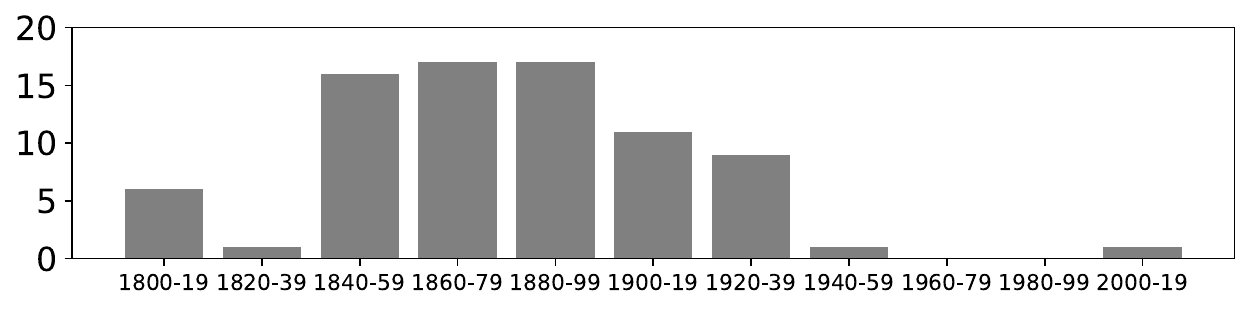}%
        } 
        \caption{\label{fig:novel-metadata} Distribution of publication dates for novels used in our main experiment. Most of our novels were published between the Victorian Era (1837--1901) and the start of World War II (1939--1945).}
    \end{figure}

    The texts used in our study consist of the 100 most downloaded books from Project Gutenberg as of August 25, 2023.\footnote{\url{https://www.gutenberg.org/ebooks/search/?sort\_order=downloads}} All texts are novels written in or translated into English. After filtering out novels published before 1800, our final dataset consists of 79 novels. As shown in \autoref{fig:novel-metadata}, most novels in our dataset were published between 1840 and 1940, roughly coinciding with the Victorian Era and the start of World War II.

    \subsection{Experimental Setup}
    
    Question Q2 introduces two independent variables: \textit{author gender} and \textit{narrator gender}. Intuitively, we expect that a novel is more likely to exhibit female objectification if it is written from a male perspective. To test this, we assume that a novel takes on a male perspective if it is written by a male author, or if it uses a male first-person narrator.

    \begin{table}
        \centering
        \footnotesize
        \begin{tabular}{l | c c c | c}
            \toprule 
            \multirow{2}{*}{\textbf{Narrator}} & \multicolumn{3}{c|}{\textbf{Author}} & \multirow{2}{*}{\textbf{Total}} \\
            & \textbf{F} & \textbf{M} & \textbf{Unknown} & \\\midrule
            1p-F & 7 & 2 & 0 & 9 \\
            1p-M & 2 & 19 & 1 & 22 \\
            3p & 13 & 31 & 1 & 45 \\
            Multiple & 1 & 2 & 0 & 3 \\\midrule 
            \textbf{Total} & 23 & 54 & 2 & 79 \\\bottomrule
        \end{tabular}
        \caption{\label{tab:metadata} The distribution of author genders and narrative perspectives in our dataset. Narrators may be third-person (3p) or first-person (1p-F or 1p-M). ``Multiple'' refers to novels with more than one narrator.}
    \end{table}
    
    \autoref{tab:metadata} shows the distribution of author and narrator features among our novels. We distinguish among three \textit{narrative perspectives}: female first-person (1p-F), male first-person (1p-M), and third-person (3p). Since the gender of third-person narrators is often unspecified, our analysis of narrator gender is limited to the comparison between 1p-F and 1p-M narrators.

    \paragraph{Hypothesis Testing.} We use a one-sample $t$-test to determine whether systematic bias has been detected in our dataset. We test the null hypothesis that the mean agency or appearance bias of our novels is 0, and we conclude that systematic bias exists if (1) the observed mean values of both bias metrics are positive, and (2) the null hypothesis is rejected with $p < \text{.05}$ for both bias metrics.

    \section{Results}
    \label{sec:results}

    \begin{table}
        \centering
        \footnotesize
        \begin{tabular}{l r r r r}
            \toprule
            & \multicolumn{2}{c}{\textbf{Agency Bias}} & \multicolumn{2}{c}{\textbf{Appearance Bias}} \\\midrule
            Overall & \textbf{.067} & ($p < \text{.001}$) & \textbf{.176} & ($p = \text{.005}$) \\\midrule
            \textit{Authors} \\
            \hspace{1em}F & .014 & ($p = \text{.660}$) & .185 & ($p = \text{.267}$) \\
            \hspace{1em}M &\textbf{.090} & ($p < \text{.001}$) & \textbf{.164}& ($p = \text{.004}$)\\\midrule 
            \textit{Narrators} \\
            \hspace{1em}1p-F & .095 & ($p = \text{.135}$) & .069 & ($p = \text{.764}$) \\ 
            \hspace{1em}1p-M & \textbf{.144} & ($p < \text{.001}$)& \textbf{.186} & ($p = \text{.015}$) \\\bottomrule
        \end{tabular}
        \caption{\label{table:mainresults} Mean agency and appearance bias scores measured in our novels dataset. Bolded results indicate evidence of systematic female objectification ($\text{mean} > \text{0}$, $p < \text{.05}$).}
    \end{table}
    
    \begin{figure}[t]
        \centering
        \resizebox{\columnwidth}{!}{%
            \includegraphics{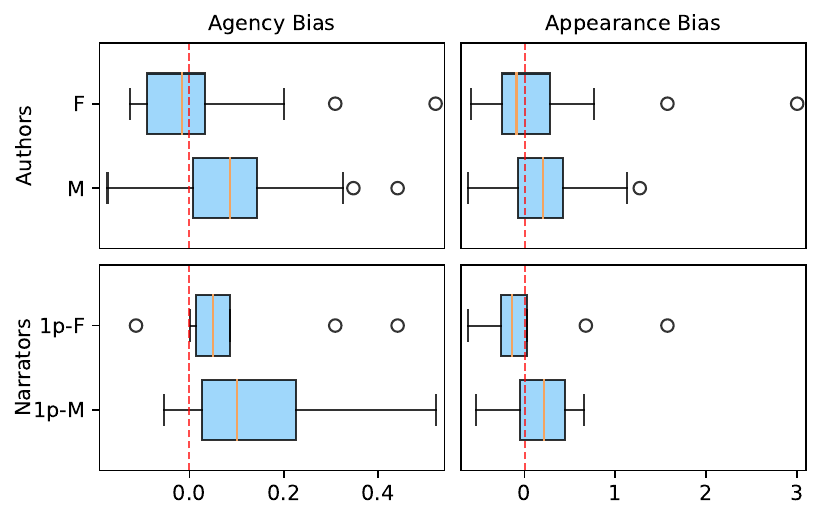}%
        } 
        \caption{\label{fig:boxplots} The distribution of agency bias and appearance bias scores for our novels, conditioned on author gender (F vs.\ M) and narrator gender (1p-F vs.\ 1p-M).}
    \end{figure}

    \begin{figure}[t]
        \centering
        \resizebox{\columnwidth}{!}{%
            \includegraphics{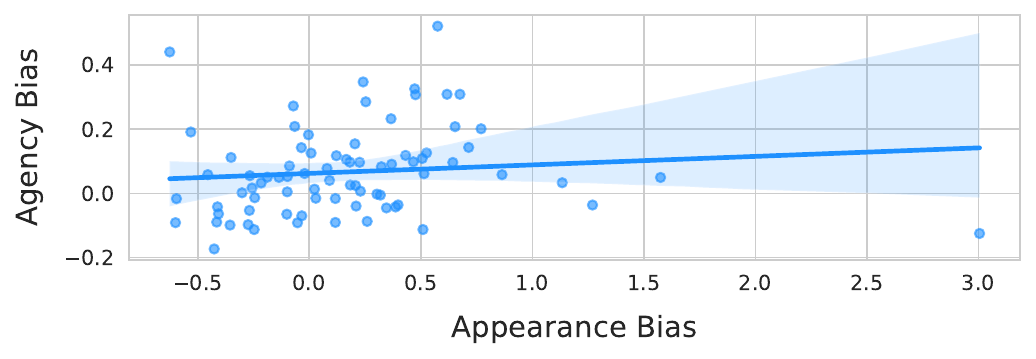}%
        } 
        \caption{\label{fig:correlation} Our two bias metrics are weakly correlated across our novels dataset ($\rho = \text{.104}$).}
    \end{figure}
    
    Our results are shown in \autoref{table:mainresults}. Both research questions are answered in the affirmative: our full set of novels shows evidence of systematic female objectification, but novels written from a female perspective---either by a female author, or using a 1p-F narrator---do not. As illustrated in \autoref{fig:correlation}, higher agency bias is associated with higher appearance bias in general, though the correlation between the two metrics is weak.
    
    The overall presence of female objectification across the entire dataset is attributable to the fact that most novels in our dataset are written from a male perspective. Although female-perspective novels have positive mean bias scores, \autoref{fig:boxplots} shows that this is driven by a small number of outliers. The majority of female-perspective novels actually exhibit \textit{negative} bias scores, with the exception that novels with a 1p-F narrator mostly exhibit positive agency bias. %

    \begin{figure}[t]
        \resizebox{\columnwidth}{!}{%
            \includegraphics{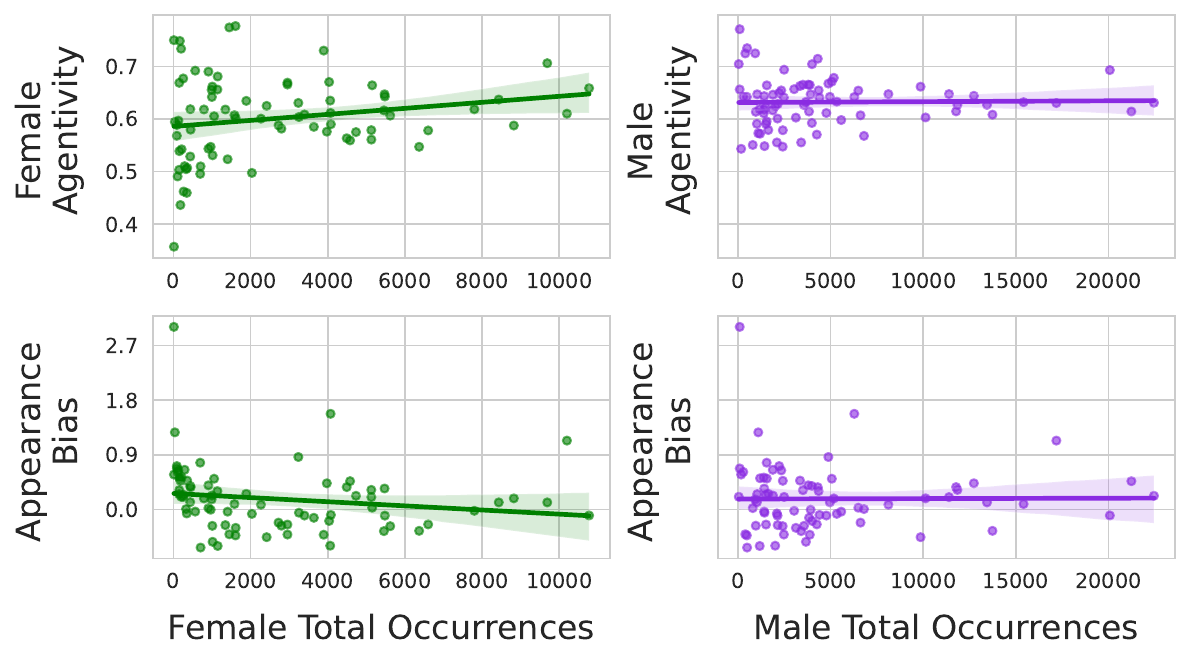}%
        } 
        \caption{\label{fig:total_cccurrence_agentivity_appearance_bias} Novels with more mentions of female characters (x-axis) exhibit higher female agentivity and lower appearance bias. }
    \end{figure}

    \autoref{fig:total_cccurrence_agentivity_appearance_bias} illustrates an asymmetry between female and male characters in terms of their contributions to agency bias and appearance bias. Although novels with more mentions of female characters exhibit higher levels of female agentivity and lower levels of appearance bias, no such relationship exists between the number of mentions of male characters and our bias metrics. This suggests that the majority of female characters are objectified, while only the most important female characters assume male levels of agency and non-appearance-related properties. In contrast, even minor male characters that occur relatively infrequently assume high levels of agency, with little attention paid to their appearance.

    \section{Related Work}

    \paragraph{Gender and Agentivity.} Gender asymmetries in grammatical agentivity have been well-documented in text corpora. \citet{macaulayDonTouchMy1997} report statistics about the agentivity of gendered entities in example sentences from a linguistics textbook. Using their results, we compute an agency bias of 1.012, almost twice as high as the highest agency bias measured in our novels dataset (.521). A similar study of linguistics journal articles \citep{kotekGenderRepresentationLinguistic2020} yields a much less extreme agency bias of .067, and \citet{dacunhaObjectifyingWomenSyntactic2022} report similar results about grammatical subjecthood in English and French corpora. Another line of work has analyzed the kinds of events that agents initiate in films \citep{rashkinEvent2MindCommonsenseInference2018}, Wikipedia entries \citep{sunMenAreElected2021}, and fairy tales \citep{toroisazaAreFairyTales2023}. Using methods similar to ours, these studies find that female agents are more likely to initiate events related to family, appearance, and sexuality, while male agents are more likely to initiate events related to work and violence.

    \paragraph{Word Embeddings and Culture.} Word embeddings are often used as summaries of cultural knowledge captured by a text corpus. \citet{hamiltonDiachronicWordEmbeddings2016}, for example, use embedding spaces trained on corpora from different time periods to track semantic change. \citet{caliskanSemanticsDerivedAutomatically2017} and \citet{gargWordEmbeddingsQuantify2018} show that word embedding spaces capture psychologically verified gender and racial stereotypes as well as disparities in demographics and labor statistics. An application of these methods to literature is presented by \citet{adukiaTalesTropesGender2022}, who show that children's books associate female characters with traits related to appearance and family relations.
 
    \paragraph{Bias and Objectification.} Surveys have found that most papers on NLP gender bias are not grounded in any explicit theory of gender \citep{devinneyTheoriesGenderNLP2022} or bias \citep{blodgettLanguageTechnologyPower2020}, though much work has been done on gender stereotypes in NLP models. Female objectification is, however, featured hate speech detection benchmarks (e.g., \citealp{fersiniSemEval2022TaskMultimedia2022}). %

    \section{Conclusion}

    This paper has developed a quantitative framework for studying female objectification in text corpora, based on our agency bias and appearance bias metrics. Our analysis of 19th and 20th century novels has found evidence of systematic female objectification, driven by the fact that popular novels from this time period are mostly written from a male perspective. Although many of the novels in our dataset do exhibit negative agency bias and appearance bias, our aggregate results suggest that female objectification is the cultural norm for English-language novels of this time period, featured in the majority of commonly downloaded male-perspective novels as well as a large minority of female-perspective novels. Our examination of frequency effects on agentivity and appearance bias suggests that agency, individuality, and subjectivity are reserved for the most important female characters, even though these attributes are readily available to male characters, both major and minor.

    Female objectification is not limited to literature or the arts. A possible application of our methods in NLP is the evaluation of female objectification in generated text, pre-training corpora, or NLP research papers. For example, agency and appearance bias scores could be reported in data cards \citep{mitchellModelCardsModel2019,gebruDatasheetsDatasets2021}. We explore such possibilities in future work.

    \section*{Acknowledgments}

    We thank the reviewers for their feedback on this paper.

    \section*{Limitations}

    \paragraph{Quality of NLP Tools.} Accurate estimation of agency bias using our analysis pipeline requires access to high-quality named entity recognizers, semantic role labelers, and gender classifiers. Agency bias results may be sensitive to differences between model instances for any of these components.

    \paragraph{Embedding Space Fine-Tuning Epochs.} The fine-tuning of embedding spaces for appearance bias calculation is hyperparameterized by the number of epochs to fine-tune for. In our study, the number of epochs was chosen in order to ensure that word embeddings are fine-tuned for the same number of training steps across novels. In other settings, appearance bias scores may not be comparable if they are computed using embeddings that have undergone different amounts of fine-tuning.

    \paragraph{Theory of Gender.} In this study, we have assumed a binary, cisnormative theory of gender. We make this assumption because, to our knowledge, none of the novels in our dataset feature a non-cisgender author or character (though there are instances of authors and characters of unknown gender). Although the definitions of agency and appearance bias do not assume cisnormativity, they do assume binarity of gender. This is a limitation of gender bias metrics more generally, since most bias metrics are designed to capture comparisons between two groups.

    \section*{Ethical Considerations}

    We do not foresee any ethical issues arising from this study.

\bibliography{custom}
\bibliographystyle{acl_natbib}

\appendix
    \section{Implementation Details}
    \label{sec:appendix}

    \begin{figure}
        \resizebox{\columnwidth}{!}{%
            \includegraphics{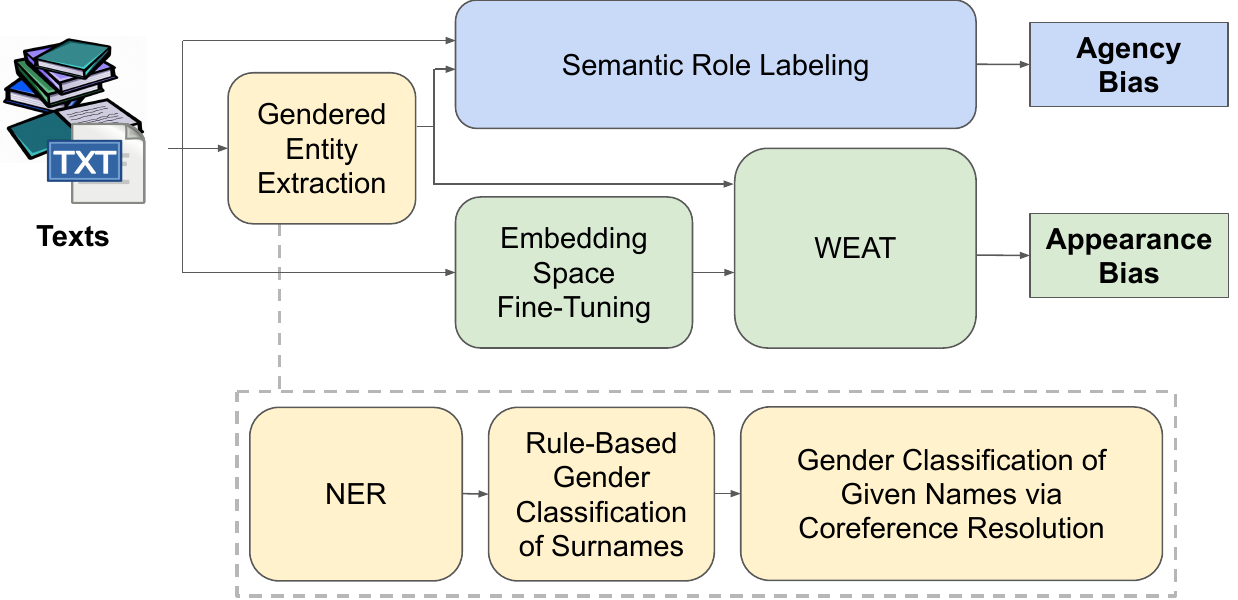}%
        } 
        \caption{\label{fig:methods_flowchart} Pipeline for calculating the agency bias and appearance bias of a text.}
    \end{figure}  

    This appendix describes the pipeline, illustrated in \autoref{fig:methods_flowchart}, that was used to calculate the agency bias and appearance bias of novels for the main experiment described in \autoref{sec:setup} and \autoref{sec:results}.

    \subsection{Gendered Entity Extraction}

    Both bias metrics rely on a \textit{gendered entity extraction} procedure that consists of the entity extraction and gender classification steps described in \autoref{sec:agencybias-calculation} and \autoref{sec:appearancebias-calculation}. For each text, this procedure produces an initial set of gendered entities appearing in the text. These entities, among others, are used in the calculation of agency bias and appearance bias.

    \paragraph{Entity Extraction.} The entity extraction step is implemented using spaCy's named entity recognizer.\footnote{\url{https://spacy.io/models/en\#en_core_web_sm}} We extract all \texttt{PERSON} entities that appear in the text at least three times.

    \begin{table}
        \centering
        \footnotesize
        \begin{tabular}{l | l}
            \toprule 
            \textbf{Female} & Madam, Madame, Mademoiselle,    \\
            \textbf{Honorifics} & Miss, Mlle, Mme, Mrs. \\\midrule
            \textbf{Male} & M., Monsieur, Mr., Sir  \\
            \textbf{Honorifics} &  \\\bottomrule
        \end{tabular}
        \caption{Titles used in the honorific heuristic.}
        \label{table:honorifics}
    \end{table}
 
    \paragraph{Gender Classification.} After the entity extraction step, we assign a gender label to each of the extracted entities using a method similar to the one used in \citet{toroisazaAreFairyTales2023}. Our gender classification procedure consists of two heuristic steps.
    \begin{itemize}
        \item \textbf{Honorific Heuristic:} Entities preceded by one of the \textit{gendered honorific titles} in \autoref{table:honorifics} are assumed to be ``surnames'' and assigned a gender label according to their honorific.
        \item \textbf{Coreference Heuristic:} We use \citeauthor{leeHigherOrderCoreferenceResolution2018}'s (\citeyear{leeHigherOrderCoreferenceResolution2018}) coreference resolution model to identify third-person pronouns co-referring with each entity that was not assigned a gender via the honorific heuristic. We label an entity as ``female'' if the model identifies more instances of \textit{she}/\textit{her}/\textit{herself} as co-referents than \textit{he}/\textit{him}/\textit{himself}, and \textit{vice versa}.
    \end{itemize}
    Our gender classification procedure attains an accuracy of 98.2\% on a manually labeled validation set consisting from 10 novels from our overall dataset. Entities that could not be assigned a gender label are excluded from the agency bias and appearance bias calculations. 

    \subsection{Agency Bias Calculation}

    Our implementation of the agency bias metric uses \citeauthor{shiSimpleBERTModels2019}'s (\citeyear{shiSimpleBERTModels2019}) semantic role labeler (SRL)\footnote{ \url{https://docs.allennlp.org/models/main/models/structured_prediction/models/srl_bert/}} to determine whether gendered entities are agents or patients. This SRL model is a tagging model, which takes a sentence as input and assigns semantic role labels to spans of tokens. 

    \paragraph{Gendered Argument Extraction.} Entities with semantic roles are collectively known as \textit{arguments}. Our agency bias calculation is based on a set of \textit{gendered arguments} extracted as follows. A span of tokens is considered a gendered argument if it has been assigned the label \texttt{ARG0} (agent) or \texttt{ARG1} (patient) by the SRL model, and one of the following conditions holds.
    \begin{enumerate}[label=(\alph*)]
        \item The span exactly matches one of the entities extracted and gender-classified during the gendered entity extraction step.
        \item The span exactly matches one of the ``common gendered entities'' appearing in \autoref{table:srl-words}.
        \item The last word of the span satisfies condition (a) or (b), and all words in the span satisfying condition (a) or (b) have the same gender label.
        \item The span contains at least one word satisfying condition (a) or (b), the last word of the span is a surname identified using the honorific heuristic during gender classification, and all words in the span satisfying condition (a) or (b) have the same gender label.
    \end{enumerate}

    \paragraph{Agentivity Calculation.} Female agentivity is calculated as follows:
    \[
    \text{female agentivity} = \frac{\text{\# of female agents}}{\text{\# of female arguments}}
    \]
    and male agentivity is calculated analogously.

    \begin{table*}
        \centering
        \footnotesize
        \begin{tabular}{l | l}
            \toprule 
            \textbf{Common} & abbess, aunt, bachelorette, baroness, bride, countess, dame, daughter, doe, druidess, duchess, \\
            \textbf{Female} & empress, female, females, firewoman, girl, girlfriend, girls, goddaughter, godmother, grandmother, \\
            \textbf{Entities} & heiress, her, heroine, herself, ladies, lady, madam, mademoiselle, mailwoman, matriarch, miss, \\
            & miss., mother, mothers, mrs, mrs., niece, nun, policewoman, princess, queen, saleswoman, she, \\
            & sister, sorceress, stepmother, widow, wife, witch, wives, woman, women \\\midrule
            \textbf{Common} & abbot, bachelor, baron, boy, boyfriend, boys, brother, druid, duke, earl, emperor, father, fathers, \\
            \textbf{Male} & fireman, friar, gentleman, godfather, godson, grandfather, groom, he, heir, him, himself, husband, \\
            \textbf{Entities} & husbands, king, knight, mailman, male, males, man, men, mister, monsieur, mr, mr., nephew, \\
            & patriarch, policeman, prince, salesman, sir, son, sorcerer, stag, stepfather, uncle, widower, wizard \\\bottomrule
        \end{tabular}
        \caption{``Common gendered entities'' that are included in the calculation of agency bias, regardless of whether they were extracted during the gendered entity extraction step.}
        \label{table:srl-words}
    \end{table*}

    \subsection{Appearance Bias Calculation}

    In addition to the male and female words appearing in \autoref{table:weat-words}, our WEAT scores also include novel-specific named entities. Since WEAT is a measure of stereotypical associations between embedding vectors, the appearance bias calculation can only include individual vocabulary items, taken out of context. These vocabulary items include all single-token entities from the gendered entity extraction step that are assigned a consistent gender label throughout the novel. Certain two-token entities are also represented in the appearance bias score, according to the following rules.
    \begin{itemize}
        \item If the first token is an honorific title appearing in \autoref{table:honorifics}, then the second token is included if it never appears after an honorific of the opposite gender label. 
        \item If the first token is not an honorific title appearing in \autoref{table:honorifics}, then the first token is included if the second item is \textit{not} a surname that has appeared with both a male and a female honorific.
    \end{itemize}

\end{document}